\documentclass[letterpaper]{article} % DO NOT CHANGE THIS
\usepackage{aaai2026}  % DO NOT CHANGE THIS
\usepackage{times}  % DO NOT CHANGE THIS
\usepackage{helvet}  % DO NOT CHANGE THIS
\usepackage{courier}  % DO NOT CHANGE THIS
\usepackage[hyphens]{url}  % DO NOT CHANGE THIS
\usepackage{graphicx} % DO NOT CHANGE THIS
\usepackage{natbib}  % DO NOT CHANGE THIS AND DO NOT ADD ANY OPTIONS TO IT
\usepackage{caption} % DO NOT CHANGE THIS AND DO NOT ADD ANY OPTIONS TO IT
\usepackage{algorithm}
\usepackage{algorithmic}
\usepackage[utf8]{inputenc} % allow utf-8 input
\usepackage[T1]{fontenc}    % use 8-bit T1 fonts

\usepackage{url}            % simple URL typesetting
\usepackage{booktabs}       % professional-quality tables
\usepackage{amsfonts}       % blackboard math symbols
\usepackage{nicefrac}       % compact symbols for 1/2, etc.
\usepackage{microtype}      % microtypography
\usepackage{xcolor}         % colors
\usepackage{graphicx}
\usepackage[utf8]{inputenc} % allow utf-8 input
\usepackage[T1]{fontenc}    % use 8-bit T1 fonts

\usepackage{url}            % simple URL typesetting
\usepackage{booktabs}       % professional-quality tables
\usepackage{amsfonts}       % blackboard math symbols
\usepackage{nicefrac}       % compact symbols for 1/2, etc.
\usepackage{microtype}      % microtypography
\usepackage{xcolor}         % colors
\usepackage{svg}
\usepackage{amsmath}
\usepackage{multirow}
\usepackage{array}
\usepackage{colortbl}
\usepackage{quoting}
\usepackage{subcaption}
\quotingsetup{leftmargin=1cm}
\usepackage{enumitem}
\usepackage{algorithm}
\usepackage{verbatim}
\usepackage{fancyvrb}
\usepackage{graphicx}
\usepackage{adjustbox}
\usepackage{algorithmic}
\usepackage{listings}
\usepackage{xcolor}
\usepackage{fvextra}
\DefineVerbatimEnvironment{MyVerbatim}{Verbatim}{breaklines, breaksymbolleft={}}

\usepackage{tikz}
\usepackage{xspace}

\lstdefinestyle{j2style}{
    basicstyle=\ttfamily\small,
    breaklines=true,
    frame=single,
    tabsize=2,
    columns=fullflexible,
    backgroundcolor=\color{gray!5}
}

\newcommand{\Lattice}{\textsc{Lattice}\xspace}

\title{Lattice: Generative Guardrails for Conversational Agents}
\usepackage{newfloat}
\usepackage{listings}
\DeclareCaptionStyle{ruled}{labelfont=normalfont,labelsep=colon,strut=off} % DO NOT CHANGE THIS
\floatstyle{ruled}
\newfloat{listing}{tb}{lst}{}
\floatname{listing}{Listing}
\author {
    Emily Broadhurst,
    Tawab Safi,
    Joseph Edell,
    Vashisht Ganesh,
    Karime Maamari
}
\affiliations {
    Distyl AI\\
    \{emily, tawab, joseph, vashisht, karime\}@distyl.ai
}

\usepackage{bibentry}
\begin{document}

\maketitle

\begin{abstract}
Conversational AI systems require guardrails to prevent harmful outputs, yet existing approaches use static rules that cannot adapt to new threats or deployment contexts. We introduce \Lattice, a framework for self-constructing and continuously improving guardrails.
% \footnote{Code, data, and experimental results will be made publicly available at \url{https://github.com/[anonymous-for-review]} upon publication.} 
\Lattice operates in two stages: construction builds initial guardrails from labeled examples through iterative simulation and optimization; continuous improvement autonomously adapts deployed guardrails through risk assessment, adversarial testing, and consolidation. Evaluated on the ProsocialDialog dataset, \Lattice achieves 91\% F1 on held-out data, outperforming keyword baselines by 43pp, LlamaGuard by 25pp, and NeMo by 4pp. The continuous improvement stage achieves 7pp F1 improvement on cross-domain data through closed-loop optimization. Our framework shows that effective guardrails can be self-constructed through iterative optimization.
\end{abstract}

\section{Introduction}

\begin{figure}[!ht]
\includegraphics[width=0.47\textwidth]{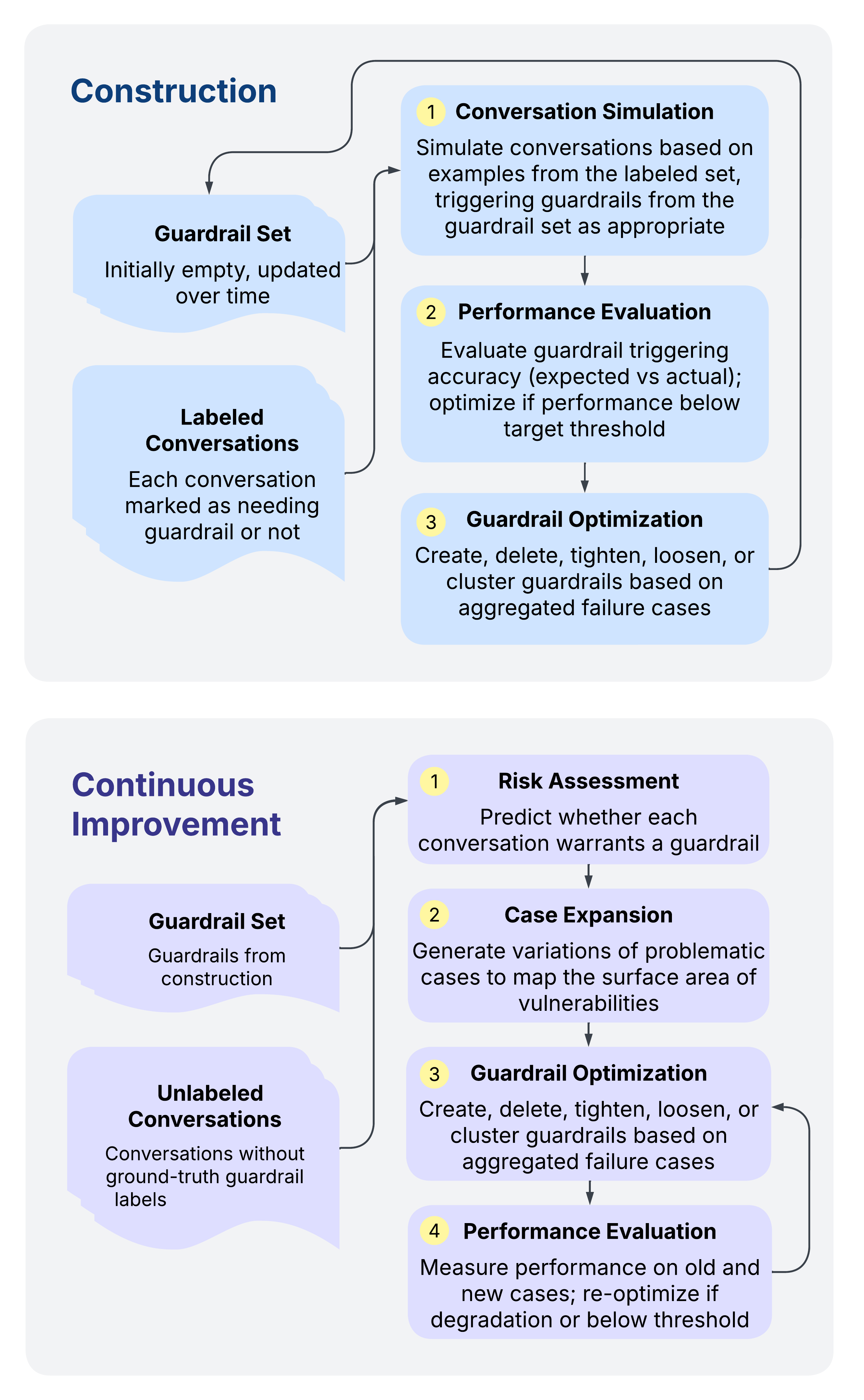}
\caption{\textbf{Two-stage framework architecture.} \textit{Construction stage} (top) generates initial guardrails from labeled conversations through three iterative steps: (1) Conversation Simulation tests current guardrails on synthetic dialogues; (2) Performance Evaluation computes precision, recall, and F1; (3) Guardrail Optimization creates, deletes, tightens, loosens, or clusters guardrails based on false positives and false negatives. \textit{Continuous improvement stage} (bottom) adapts deployed guardrails to unlabeled conversations through four steps: (1) Risk Assessment identifies coverage gaps via dual-check evaluation; (2) Case Expansion generates adversarial variations; (3) Guardrail Optimization updates policies; (4) Performance Evaluation validates changes and reverts if performance degrades.}
\label{fig:constructor}
\end{figure}

The deployment of large language models in conversational AI systems presents a fundamental tension between capability and safety. Although these systems must engage naturally in diverse contexts, they also require mechanisms to prevent harmful outputs in real-world deployments~\cite{ayyamperumal2024currentstatellmrisks,hakim2024needguardrailslargelanguage, 10.1145/3644815.3644958}.

Current approaches to conversational safety are predominantly based on static guardrail mechanisms~\cite{dong2024buildingguardrailslargelanguage,rebedea2023nemoguardrailstoolkitcontrollable}. These systems employ fixed rule sets designed to filter input queries and LLM outputs based on predetermined patterns. However, static guardrails face two critical limitations that compromise their effectiveness in deployed systems. First, they cannot adapt to attack vectors or conversational contexts that emerge post-deployment~\cite{yang2024benchmarkingllmguardrailshandling}. Second, they exhibit brittleness when faced with adversarial inputs specifically crafted to circumvent existing protections~\cite{goyal2024llmguardguardingunsafellm}.

These limitations point to a deeper algorithmic challenge: \emph{how can guardrail systems self-evolve to address emergent threats while maintaining safety coverage?} This requires moving beyond static guardrail application toward systems capable of learning, identifying coverage gaps, and refining protection mechanisms through experiences.

We introduce \Lattice, a framework that treats guardrail construction and adaptation as continuous optimization problems. Rather than relying on fixed rules defined at design time, \Lattice constructs initial guardrail sets through iterative conversation simulation, evaluating candidate policies against labeled training data and refining them based on observed failure modes. After deployment, \Lattice continuously improves these guardrails by monitoring conversations for coverage gaps, expanding edge cases through adversarial beam search, and updating policies when performance degrades. Evaluated on ProsocialDialog~\cite{kim2022prosocialdialog}, this approach outperforms static baseline systems and demonstrates cross-domain adaptation.

Our primary contributions are: (1) A simulation-based construction method that learns compact guardrail sets from minimal labeled data through iterative optimization, outperforming manually designed static systems. (2) A closed-loop continuous improvement system combining dual-check risk assessment, adversarial case expansion, and policy optimization that enhances deployed guardrails, achieving measurable performance gains in cross-domain settings.

\section{Related Work}

\subsection{Static safety mechanisms at training and inference time}
Safety mechanisms for language models are typically introduced either during training or at inference time. Training time methods include safety fine tuning and guardrail aware adaptation~\cite{kumar2024finetuningquantizationllmsnavigating}, as well as parameter efficient approaches such as LoRA based safety modules~\cite{fomenko2024notelora,hsu2025safelorasilverlining}. These methods embed constraints directly into model parameters, which can yield strong performance within the training distribution. However, they require new optimization cycles whenever threats change, limiting rapid adaptation to novel jailbreak patterns~\cite{huang2025virusharmfulfinetuningattack}. 

Inference time guardrail systems, including NeMo Guardrails~\cite{rebedea2023nemoguardrailstoolkitcontrollable} and LlamaGuard~\cite{fedorov2024llamaguard31bint4compact}, represent the prevailing runtime strategy. These systems achieve broad coverage of predefined safety categories but their rules remain fixed at design time, which constrains their ability to handle multi-turn jailbreaks or emerging violation types~\cite{yang2024benchmarkingllmguardrailshandling,yu2024cosafeevaluatinglargelanguage}. \Lattice builds on this line of work by introducing an inference time mechanism that not only applies rules but also generates, edits, and consolidates them through a continuous optimization process.

\subsection{Programmed guardrails and moderation models}
Rule based toolkits such as NeMo Guardrails provide programmable rails, dialogue flows, and domain specific policy enforcement~\cite{rebedea2023nemoguardrailstoolkitcontrollable}. Moderation models such as LlamaGuard~\cite{fedorov2024llamaguard31bint4compact} and ShieldGemma~\cite{zeng2024shieldgemmagenerativeaicontent} classify content categories for safety filtering. These systems are effective for predefined risks, yet studies highlight gaps in coverage for paraphrased intent, cross category drift, and extended dialogue context~\cite{yang2024benchmarkingllmguardrailshandling,yu2024cosafeevaluatinglargelanguage}. Our approach uses such systems as components inside a broader feedback loop: general purpose classifiers serve as baseline evaluators, while \Lattice induces new guardrails to address their blind spots.

\subsection{Adversarial robustness, red teaming, and multi turn jailbreaks}
A substantial literature explores adversarial prompting and jailbreak attacks on safety systems. Red teaming frameworks demonstrate that carefully staged inputs can bypass filters and that multi turn dialogues can erode safety boundaries by gradually shifting context~\cite{goyal2024llmguardguardingunsafellm}. Multi step evaluations reveal that contextual drift, latent goal pursuit, and role play exploitation often lead to safety failures~\cite{yang2024benchmarkingllmguardrailshandling}. HAICOSYSTEM~\cite{zhou2024haicosystemecosystemsandboxingsafety} introduces sandboxed evaluation across diverse domains, emphasizing large scale testing of system vulnerabilities. These works focus primarily on identifying weaknesses. \Lattice complements them by integrating adversarial testing with automatic policy improvement, converting discovered jailbreaks into new guardrail rules without manual patching.

\subsection{Learning based refinement and critique driven methods}
Iterative learning approaches use feedback loops to refine model behavior. Self Refine~\cite{madaan2023selfrefineiterativerefinementselffeedback} and critique driven refinement~\cite{ke-etal-2024-critiquellm,ye2024improvingrewardmodelssynthetic} demonstrate that self feedback can improve textual quality and consistency, though their focus lies on generation fidelity rather than safety. Reward model scaling studies~\cite{rafailov2024scalinglawsrewardmodel} show that feedback driven alignment improves safety and helpfulness, but these methods often rely on thousands of human annotated feedback samples per iteration and may encounter reward hacking effects~\cite{yan2024rewardrobustrlhfllms}. \Lattice differs by employing synthetic feedback derived from observed guardrail failures and adversarial expansions, enabling unsupervised improvement.

\section{Methodology}

\subsection{Construction Stage}
The construction stage learns an initial guardrail set from labeled conversations via iterative refinement (Figure~\ref{fig:constructor}, Algorithm~\ref{alg:construction}). Each iteration comprises three steps that (i) expand behavioral coverage, (ii) evaluate detection performance, and (iii) apply targeted edits to the guardrail set while enforcing an acceptance criterion that prevents regressions.

\subsubsection{Conversation Simulation}
\textit{(Algorithm~\ref{alg:construction}, lines 7--11)}\;
For each labeled conversation $(c_i, y_i)$, the system samples a user persona and simulates a multi-turn dialogue conditioned on the current guardrails $R$. We then evaluate whether any $r \in R$ triggers on the simulated transcript $s$, appending $(c_i, y_i, \texttt{triggered})$ to a validation set $V$. This procedure exposes $R$ to diverse conversational dynamics and yields observed trigger patterns beyond the original labels, increasing the effective variation encountered during optimization.

\subsubsection{Performance Evaluation}
\textit{(Algorithm~\ref{alg:construction}, lines 12--20)}\;
From $V$, we compute $\text{TP}, \text{FP}, \text{FN}, \text{TN}$ by comparing $y_i$ with the observed trigger outcome and report precision $P=\frac{\text{TP}}{\text{TP}+\text{FP}}$, recall $R=\frac{\text{TP}}{\text{TP}+\text{FN}}$, and $F_1=\frac{2PR}{P+R}$ as the primary selection metric. This diagnostic decomposition identifies whether error is dominated by over-sensitivity (FP) or under-coverage (FN), providing sufficient statistics to guide the subsequent optimization step. If $F < F^{\star}$ and a prior best configuration $R^{\star}$ exists, we revert to $R^{\star}$; if $F \geq F^{\star}$, we promote the current configuration to $R^{\star}$ and update $F^{\star}$. Early stopping is triggered when $F \geq \tau$, where $\tau$ is a user-specified threshold.

\subsubsection{Guardrail Optimization}
\textit{(Algorithm~\ref{alg:construction}, lines 21--40)}\;
Based on the failure modes observed in $V$, we apply targeted edits to $R$ as follows:
\vspace{0.2em}
\begin{itemize}
  \item \textbf{False negatives related to existing guardrails:} broaden the guardrail involved to increase sensitivity.
  \item \textbf{False negatives unrelated to existing guardrails:} synthesize a new specialized guardrail to cover the uncovered pattern.
  \item \textbf{False positives:} refine the triggered guardrail to reduce over-flagging while preserving coverage.
  \item \textbf{Unused guardrails:} remove rules that never trigger to prevent bloat and improve interpretability.
  \item \textbf{Redundant guardrails:} optionally cluster similar rules and consolidate them into a single policy.
\end{itemize}
\vspace{0.2em}
Each proposed edit is \emph{accepted} only if it maintains or improves the best observed score, i.e., $F_{\text{new}} \geq F^{\star}$; otherwise the system \emph{reverts} to $R^{\star}$. This acceptance rule enforces monotonic, non-decreasing $F_1$ over accepted configurations, ensuring stable convergence of the construction process.

\subsection{Continuous Improvement Stage}
The continuous improvement stage adapts the deployed guardrails to the unlabeled deployment data through structured testing and optimization (Algorithm~\ref{alg:continuous}). Each iteration executes four steps --risk assessment, case expansion, guardrail optimization, and performance evaluation -- to maintain or improve safety coverage in response to novel conversational patterns.

\subsubsection{Risk Assessment}
\textit{(Algorithm~\ref{alg:continuous}, lines 9--16)}\;
Each new conversation $u$ undergoes a dual-evaluation procedure: the system tests both (i) current specialized guardrails $R$ and (ii) a general-purpose safety guardrail $r_{\text{general}}$. The dual-check mechanism identifies coverage gaps by comparing activation states; cases where $r_{\text{general}}(u)=1$ and $\bigvee_{r\in R}r(u)=0$ indicate \emph{missed threats-instances}---instances where the general model detects risk that the specialized set fails to capture. The resulting set of discrepancies, denoted $G$, serves as the seed for a subsequent adversarial exploration.

\subsubsection{Case Expansion}
\textit{(Algorithm~\ref{alg:continuous}, lines 17--26)}\;
For each $u \in G$, the system performs multi-turn adversarial search to generate conversation variants that investigate known and potential weaknesses. An \emph{attacker model} $M_a$ produces candidate user turns designed to bypass existing guardrails while preserving the inferred harmful intent; a \emph{target model} $M_t$ responds under the current guardrail policy. The beam search with configurable width $k$ and depth $d$ expands the conversation tree, producing a set of labeled leaf nodes $\mathcal{E}$ categorized as:
\vspace{0.2em}
\begin{itemize}
  \item \emph{successful attacks} (guardrail bypasses)
  \item \emph{blocked attacks} (guardrail triggered correctly)
  \item \emph{false alarms} (benign content flagged in error)
\end{itemize}
\vspace{0.2em}
This step operationalizes adversarial testing, generating diverse challenge cases.

\subsubsection{Guardrail Optimization}
\textit{(Algorithm~\ref{alg:continuous}, lines 27--42)}\;
The system applies targeted updates to $R$ conditioned on the analysis of $\mathcal{E}$:
\vspace{0.2em}
\begin{itemize}
  \item \textbf{Successful attacks:} broaden or synthesize new guardrails to cover novel attack strategies.
  \item \textbf{False alarms:} refine corresponding guardrails to reduce over-sensitivity.
  \item \textbf{Redundant policies:} cluster semantically similar guardrails and consolidate them to prevent drift and complexity growth.
\end{itemize}
\vspace{0.2em}
Each proposed update produces a candidate set $R'$ that is validated against a held-out evaluation split. Updates are accepted only if $F_1(R') \geq F_1(R)$, enforcing non-decreasing performance across improvement cycles.

\subsubsection{Performance Evaluation}
\textit{(Algorithm~\ref{alg:continuous}, lines 43--47)}\;
Updated guardrails $R'$ are re-evaluated on both the original risk-assessment cases $G$ and the adversarially expanded cases $\mathcal{E}$. If $F_1(R') < F_1(R)$ or performance falls below a specified degradation threshold, the system reverts to the prior configuration $R$. Otherwise, $R'$ is promoted to the deployed set, thereby closing the continuous-improvement loop and enabling adaptive response to emergent threats.

\subsection{Implementation via Prompted Language Models}
All algorithmic operations in the construction and continuous improvement stages are implemented as structured prompt calls to large language models (LLMs). To ensure fair comparison with baselines and computational efficiency, all operations use \texttt{gpt-4o-mini} uniformly across the pipeline. This includes persona sampling, conversation simulation, guardrail generation and refinement, adversarial attack generation, and all testing procedures. 

Each LLM call follows a fixed prompt schema to ensure consistency and facilitate automatic parsing of results:
\begin{MyVerbatim}
### TASK
<Defines the operation's purpose and
establishes the model's role>

### INSTRUCTIONS
<Provides step-by-step procedures, 
specific requirements, and 
constraints>

### OUTPUT FORMAT
<Specifies the expected response schema, 
typically JSON with fields for
generated content and reasoning>
\end{MyVerbatim}

This standardized structure constrains model behavior, enforces reproducible output formats, and allows the system to programmatically integrate LLM outputs within the guardrail construction and improvement loops.

\section{Experimental setup}
\begin{figure}[t]
\hspace*{-0.4cm}
% \centering
\includegraphics[width=0.5\textwidth]{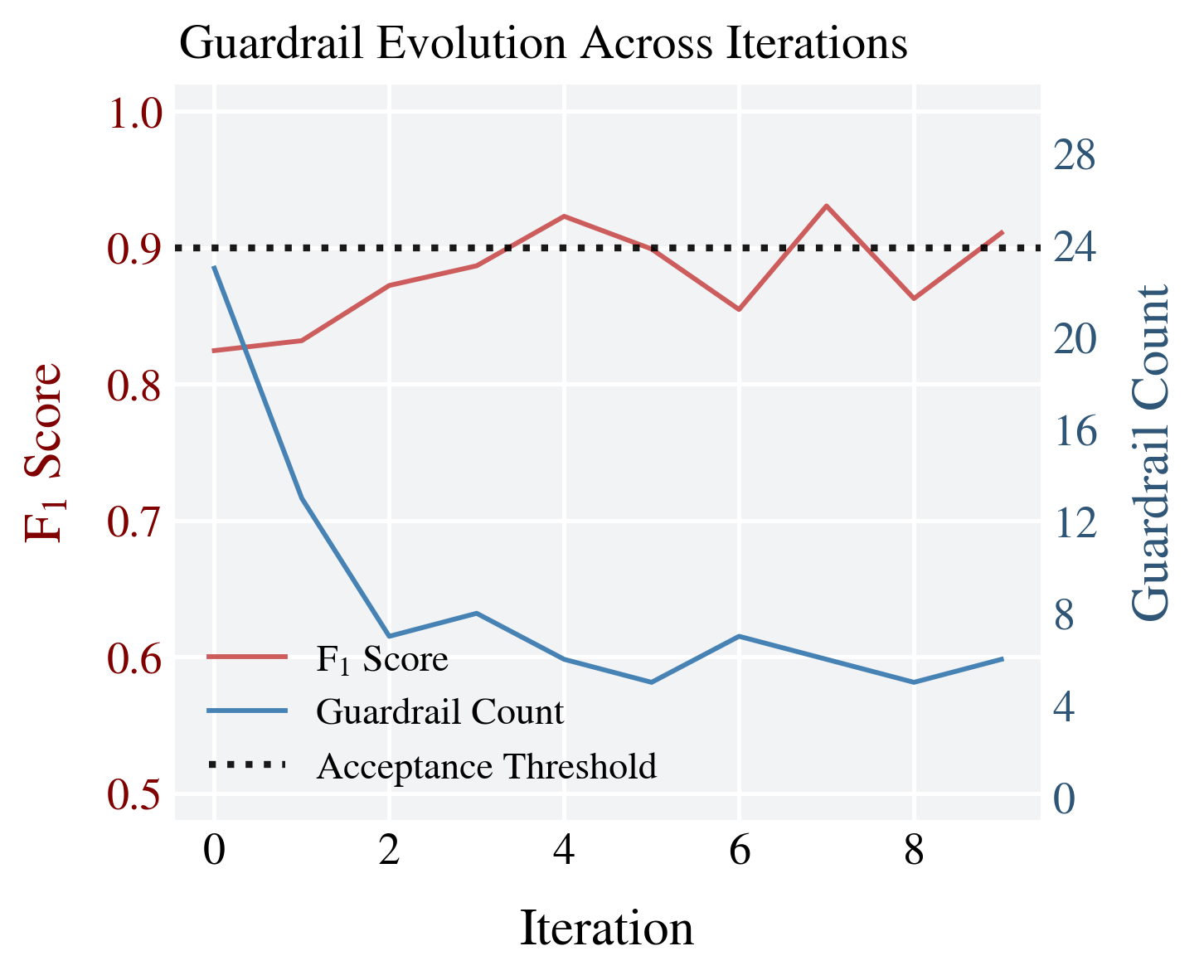}
\caption{\textbf{Iterative construction from 100 labeled examples.} F1 score (left axis, red) and guardrail count (right axis, blue) across 10 construction iterations. F1 improves from 82\% (iteration 0, 23 guardrails) to 93\% (iteration 7, 6 guardrails), exceeding the early stopping threshold ($F_1 \geq 0.90$, black dotted line) at iteration 4. Guardrail consolidation reduces the set from 23 to 6 policies through iterative clustering while maintaining performance. The system exhibits convergence with non-monotonic but improving F1 trajectory.}
\label{fig:construction_iterations}
\end{figure}

\begin{table}[t]
\centering
\begin{tabular}{ll}
\toprule
\textbf{Parameter} & \textbf{Value} \\
\midrule
Max construction iterations & 10 \\
Early stop F1 threshold & 0.90 \\
Simulated conversation turns & 3 \\
Iteration selection metric & F1 score \\
Beam width (case expansion) & 3 \\
Tree depth (case expansion) & 10 \\
Holdout evaluation runs & 5 \\
\bottomrule
\end{tabular}
\caption{\textbf{Key hyperparameters for reproducibility.} Construction iterates up to 10 times with early stopping when $F_1 \geq 0.90$. Each iteration simulates 3-turn conversations and selects updates based on F1 score comparison. Continuous improvement uses beam search with width 3 and depth 10 for adversarial case expansion.}
\label{tab:hyperparameters}
\end{table}

\subsection{Dataset}
We evaluate on \textsc{ProsocialDialog}~\cite{kim2022prosocialdialog}, a corpus of 58{,}000 multi-turn conversations between users and AI assistants annotated for conversational safety. Each dialogue is labeled as either \texttt{harmful} (requiring guardrail intervention) or \texttt{benign} (safe). To ensure consistency, we filter out ambiguous or multi-intent examples and retain only those with unambiguous binary annotations. From this filtered subset, we construct three evaluation splits: $\mathcal{D}_{\text{train}}$ (100 conversations from social domain), $\mathcal{D}_{\text{test}}$ (652 conversations from social domain, disjoint from train, representing 10\% of the social domain data), and $\mathcal{D}_{\text{improve}}$ (100 conversations from ethics domain for cross-domain testing). We treat the social and ethics domains as distinct subspaces within the corpus to better capture differences between interpersonal sensitivity and moral reasoning. Training and test sets are balanced between harmful and benign labels.

\subsection{Evaluation Metrics}
We report standard binary classification metrics. Precision ($P$) quantifies the fraction of flagged conversations that are truly harmful. Recall ($R$) measures the fraction of harmful conversations correctly identified. The F1-score provides the harmonic mean of precision and recall. All holdout and baseline results are averaged over 5 independent runs on the same test set to estimate measurement variance.

\subsection{Baseline Systems}
We compare \Lattice against three representative static guardrail baselines:
\begin{enumerate}
    \item \textit{Keyword} — deterministic keyword matching over predefined lexical patterns (e.g., \textit{suicide}, \textit{kill}, \textit{violence}, \textit{abuse}).
    \item \textit{LlamaGuard}~\cite{fedorov2024llamaguard31bint4compact} — Meta’s Llama-Guard-3-8B content moderation model accessed via the Together~AI API.
    \item \textit{NeMo}~\cite{rebedea2023nemoguardrailstoolkitcontrollable} — NVIDIA’s NeMo Guardrails framework configured with a \texttt{gpt-4o-mini} backend.
\end{enumerate}
All baselines are evaluated on the same 652-conversation test set using the same preprocessing pipeline to ensure comparability.

\subsection{Hyperparameters}
Table~\ref{tab:hyperparameters} summarizes key experimental settings. The construction stage iterates up to $T=10$ cycles with early stopping when $F_1 \geq 0.90$. Each iteration simulates three-turn dialogues, computes $F_1$, and applies targeted guardrail optimization based on precision–recall analysis. Model selection across iterations follows an $F_1$-based comparison criterion. The continuous improvement stage employs beam search for adversarial case expansion with width $k=3$ and depth $d=10$. All reported experiments are conducted under these fixed hyperparameters.

\subsection{Research Questions}
We evaluate \Lattice across three core research objectives:

\begin{enumerate}
    \item \textbf{RQ1 — Self-Construction:} Can \Lattice self-construct an effective guardrail set from a limited sample of 100 labeled conversations such that the resulting model achieves $F_1 \geq 0.90$ on unseen data?
    
    \item \textbf{RQ2 — Baseline Comparison:} Does the guardrail set produced by \Lattice outperform representative static baselines—\textit{Keyword}, \textit{LlamaGuard}, and \textit{NeMo}—in detecting harmful conversations?
    
    \item \textbf{RQ3 — Continuous Improvement:} Can the continuous improvement stage, when deployed on unlabeled data, improve $F_1$ without human supervision or manual intervention?
\end{enumerate}

These questions collectively test the extent to which \Lattice can (i) construct guardrails with minimal supervision, (ii) achieve or exceed state-of-the-art performance relative to existing static systems, and (iii) continuously adapt to emerging conversational risks in a fully automated manner.

\section{Results}\label{sec:results}

\subsection{Self-Construction (RQ1)}

Figure~\ref{fig:construction_iterations} shows construction performance across 10 iterations on 100 labeled examples. The system starts with 23 initial guardrails at iteration 0 achieving $F_1$=82\%. Through iterative refinement, $F_1$ improves to 93\% by iteration 7, surpassing the early stopping threshold ($F_1 \geq 0.90$) at iteration 4. Guardrail consolidation reduces the set from 23 to 6 policies while maintaining performance. Sample constructed guardrails demonstrating the specificity and structure of generated policies are shown in Figures~\ref{fig:guardrail_illegal} and~\ref{fig:guardrail_inappropriate}.

\begin{table}[t]
\centering
\small
\begin{tabular}{lccc}
\toprule
\textbf{Metric} & \textbf{Initial} & \textbf{Final} & \textbf{Holdout} \\
\midrule
Precision & $73\%$ & $91\%$ & $90\% \pm 1\%$ \\
Recall & $94\%$ & $96\%$ & $93\% \pm 1\%$ \\
F1 Score & $82\%$ & $93\%$ & $91\% \pm 1\%$ \\
\midrule
Guardrails generated & \multicolumn{3}{c}{23 (initial) $\rightarrow$ 6 (final)} \\
Iterations to convergence & \multicolumn{3}{c}{4 (of max 10)} \\
\bottomrule
\end{tabular}
\caption{\textbf{Self-construction performance on training and test sets.} Initial construction (iteration 0) starts with 23 guardrails achieving 82\% F1. Final construction (best iteration) improves to 93\% F1 with 6 guardrails after consolidation. Holdout evaluation over 5 independent runs yields 91\% $\pm$ 1\% F1 (mean ± 95\% CI), with the 2pp gap analyzed in Section~\ref{sec:discussion}. The system reaches the early stopping threshold ($F_1 \geq 0.90$) at iteration 4. High recall (93\%) ensures comprehensive coverage while maintaining strong precision (90\%).}
\label{tab:construction_results}
\end{table}

The construction stage achieves $F_1 = 91\% \pm 1\%$ on held-out data from 100 labeled training examples. While final construction performance (93\%) exceeds holdout performance (91\%), indicating some overfitting, the absolute holdout F1 remains substantially higher than initial construction (82\%), demonstrating effective learning. The system prioritizes recall (93\% $\pm$ 1\%), ensuring comprehensive coverage of harmful content, while maintaining strong precision (90\% $\pm$ 1\%).

\subsection{Baseline Comparison (RQ2)}

Table~\ref{tab:baselines} compares \Lattice against static guardrail systems. \Lattice achieves the highest F1 (91\%), outperforming keyword matching (48\%), LlamaGuard (66\%), and NeMo (87\%). The 43pp improvement over keyword matching demonstrates the value of learned guardrails. Compared to NeMo (4pp improvement), \Lattice achieves comparable F1 with higher precision but slightly lower recall (90\%/93\% vs 82\%/93\%). Compared to LlamaGuard (25pp improvement), \Lattice achieves substantially higher recall (93\% vs 50\%) with lower but still strong precision (90\% vs 98\%).

\begin{table}[t]
\centering
\small
\begin{tabular}{lccc}
\toprule
\textbf{System} & \textbf{Precision} & \textbf{Recall} & \textbf{F1} \\
\midrule
Keyword & $95\% \pm 0\%$ & $32\% \pm 0\%$ & $48\% \pm 0\%$ \\
LlamaGuard-8B & $98\% \pm 0\%$ & $50\% \pm 0\%$ & $66\% \pm 0\%$ \\
NeMo & $82\% \pm 1\%$ & $93\% \pm 0\%$ & $87\% \pm 0\%$ \\
\midrule
Lattice & $90\% \pm 1\%$ & $93\% \pm 1\%$ & $91\% \pm 1\%$ \\
\bottomrule
\end{tabular}
\caption{\textbf{Comparison against static guardrail baselines.} All systems evaluated on the same 652-conversation test set over 5 independent runs (balanced 50\% harmful). \Lattice achieves 91\% F1, outperforming keyword matching (48\%), LlamaGuard-8B (66\%), and NeMo Guardrails (87\%). Improvements of 43pp, 25pp, and 4pp respectively. \Lattice achieves high precision and recall (90\%/93\%) compared to LlamaGuard's high precision but low recall (98\%/50\%) and NeMo's recall-focused approach (82\%/93\%).}
\label{tab:baselines}
\end{table}

\subsection{Continuous Improvement (RQ3)}

Table~\ref{tab:continuous_stats} demonstrates automated guardrail adaptation through the continuous improvement stage. Starting from baseline performance of 86\% F1 on the improvement dataset (cross-domain ethics data), the system executes the full feedback loop. The dual-check risk assessment identifies 8 conversations where general safety classifiers trigger but specific guardrails do not, indicating potential coverage gaps. For each gap, beam search with width 3 and depth 10 generates adversarial conversation variants, producing 24 test cases that probe guardrail boundaries. The optimization component performs 5 targeted guardrail updates: broadening policies that miss related cases and creating new policies for novel patterns. The adapted guardrails achieve 93\% F1, representing a 7pp improvement over baseline, validating that the continuous improvement loop enhances safety coverage even in cross-domain settings.

% Table~\ref{tab:continuous_stats} shows adaptation through the continuous improvement stage. Starting from initial construction performance ($F_1=85\%$), the system identifies 2 coverage gaps via risk assessment and generates adversarial test cases through beam search. After guardrail optimization, final performance achieves $84\%$ F1. While the absolute improvement is modest ($-1$pp), the system successfully identifies and addresses coverage gaps without human intervention, validating the operational capability of the feedback loop.

\begin{table}[t]
\centering
\small
\begin{tabular}{lc}
\toprule
\textbf{Metric} & \textbf{Value} \\
\midrule
Coverage gaps identified & 8 \\
Adversarial tests generated & 24 \\
Guardrail updates performed & 5 \\
\midrule
Initial F1 (before) & $86\%$ \\
Final F1 (after) & $93\%$ \\
Change & $+7$pp \\
\bottomrule
\end{tabular}
\caption{\textbf{Continuous improvement operational statistics.} Starting from constructed guardrails with 86\% F1 on cross-domain data (ethics), the system identifies 8 coverage gaps via dual-check risk assessment, generates 24 adversarial test cases through beam search, and performs 5 guardrail updates. Final performance achieves 93\% F1, representing a 7pp improvement through the risk assessment, case expansion, and optimization loop.}
\label{tab:continuous_stats}
\end{table}

\subsection{Discussion}\label{sec:discussion}

Our results demonstrate that \Lattice constructs effective guardrails through iterative optimization, achieving strong holdout performance while maintaining compact policy sets. The system converges efficiently and substantially outperforms static baselines across all metrics.

\subsubsection{Precision-recall trade-offs and configurability.} Table~\ref{tab:baselines} reveals distinct precision-recall profiles across systems: LlamaGuard achieves high precision (98\%) but low recall (50\%), capturing only the most obvious violations while avoiding false positives; NeMo prioritizes recall (93\%) with lower precision (82\%), flagging more broadly at the cost of over-triggering; \Lattice achieves strong precision (90\%) and high recall (93\%). The 10\% false positive rate in \Lattice's current configuration reflects an explicit optimization toward comprehensive harmful content detection. This trade-off is appropriate for safety-critical deployments where missing harmful content (false negatives) carries greater risk than occasional over-flagging of benign conversations (false positives).

Crucially, \Lattice's precision-recall balance is \textit{configurable} rather than fixed. The framework supports two optimization modes: (1) F1-based selection (harmonic mean of precision and recall), or (2) weighted scoring with user-defined coefficients ($\alpha P + \beta R$). F1 optimization penalizes precision-recall imbalance, ensuring both metrics remain high. Weighted scoring allows asymmetric optimization: applications prioritizing user experience can set $\alpha=2.0, \beta=1.0$ to favor precision (fewer false positives), while safety-critical deployments can set $\alpha=1.0, \beta=2.0$ to favor recall (fewer false negatives). This configurability enables deployment-specific optimization without code changes, allowing organizations to align guardrail behavior with their risk tolerance and operational constraints. Our evaluation uses F1-based selection, representing a balanced default that equally penalizes precision and recall deficiencies.

\subsubsection{Generalization and overfitting analysis.} Table~\ref{tab:construction_results} shows a 2pp drop in F1 from final construction (93\%) to holdout evaluation (91\%). This gap warrants careful interpretation. On one hand, the performance decrease indicates some degree of overfitting to the 100-example training set. On the other hand, the absolute holdout performance (91\%) remains substantially higher than initial construction (82\%), representing a 9pp net improvement. Additionally, the 91\% holdout F1 significantly exceeds all static baselines, demonstrating that despite the generalization gap, the learned guardrails capture meaningful safety patterns.

The 2pp gap is within expected bounds for classification tasks trained on 100 examples. Comparing precision and recall components reveals the source: final construction achieves 91\% precision on training data, maintaining similar holdout precision (90\%), while recall decreases from 96\% to 93\%. This suggests the guardrails generalize well but are slightly more conservative on new data. The modest generalization gap, combined with strong absolute performance, indicates the approach is viable for deployment.

The continuous improvement stage demonstrates effective unsupervised adaptation. Starting from constructed guardrails with 86\% F1 on cross-domain improvement data (ethics), the system identifies 8 coverage gaps through dual-check risk assessment, generates 24 adversarial test cases via beam search exploration, and performs 5 targeted guardrail updates. The final adapted guardrails achieve 93\% F1, representing a 7pp improvement. This validates that the framework can enhance deployed guardrails through structured feedback loops, even when adapting to new domains.

\subsubsection{Computational cost analysis.} The construction stage consumes approximately 53.6 million tokens (\textasciitilde\$20 at \texttt{gpt-4o-mini} pricing) and requires 46 minutes of runtime to train on 100 labeled examples. This cost contrasts with static baseline systems—keyword matching, LlamaGuard, and NeMo Guardrails—which incur no construction overhead but require manual rule specification and cannot self-adapt post-deployment. While the upfront computational investment is substantial, it must be contextualized within deployment scale. For production conversational AI systems serving millions of users, construction costs are amortized across all subsequent interactions. Consider a customer service chatbot handling 100,000 daily conversations: a one-time cost of \$20 and 46 minutes yields guardrails protecting 36.5 million annual conversations, reducing to \$0.0000005 per protected conversation. For safety-critical domains where a single harmful output could have severe consequences, this investment is justified by superior performance over static systems (4--43pp improvement), automated construction from minimal supervision, and continuous adaptation capabilities that static systems lack. The framework's configurability further enables cost-performance trade-offs: organizations can adjust iteration counts, clustering aggressiveness, and precision-recall weights based on their risk tolerance and budget constraints.

\subsubsection{Limitations and Future Directions}

While \Lattice demonstrates effective guardrail construction and adaptation, several limitations warrant consideration. First, evaluation focuses exclusively on English-language conversations within the ProsocialDialog domain; generalization to multilingual settings, domain-specific contexts, and cross-cultural safety norms remains untested. Second, all pipeline operations use \texttt{gpt-4o-mini} uniformly; exploring additional models, particularly reasoning models, could identify optimal model-operation pairings to improve quality or reduce costs. Third, we evaluate with a fixed 100-example training set; investigating how training set size affects holdout and continuous improvement performance would inform data collection requirements. Fourth, automated refinement raises interpretability questions for regulated domains requiring audit trails.

Future work should investigate: (1) multilingual and cross-domain robustness; (2) training set size ablations to determine data efficiency bounds; (3) model selection studies comparing reasoning models against fast inference models for different pipeline operations; (4) adversarial testing against sophisticated jailbreak techniques; (5) human-in-the-loop variants for high-stakes decisions; and (6) long-term deployment studies measuring adaptation and drift.

\section{Conclusion}

We presented \Lattice, a framework enabling conversational AI systems to self-construct and continuously improve guardrails.. The framework operates through two stages: construction generates initial guardrails from labeled training examples via iterative simulation and optimization; continuous improvement adapts deployed guardrails through risk assessment, adversarial testing, and consolidation. 

Beyond accuracy, the system is configurable—trading precision vs. recall to match deployment risk—and practical, with a one-time construction cost on the order of minutes and dollars that amortizes at scale. Framing safety this way reconciles ethics-driven desiderata with engineering constraints: the guardrails self-construct, self-audit, and self-improve, enabling sustained coverage as threats and contexts shift. This is a step toward resilient, auditable safety layers that keep pace with real-world dialogue systems without proportional increases in human oversight.

\bibliography{aaai2026}

\newpage
\begin{figure*}[p]
\centering
\begin{minipage}{0.9\textwidth}
\begin{algorithm}[H]
\caption{Construction Stage: Iterative Guardrail Optimization}
\label{alg:construction}
\small
\begin{algorithmic}[1]
\STATE \textbf{Input:} Labeled dataset $\mathcal{D} = \{(c_i, y_i)\}_{i=1}^n$ where $y_i \in \{0,1\}$ indicates if conversation $c_i$ requires guardrail
\STATE \phantom{Input:} Max iterations $T$, F1 threshold $\tau$
\STATE \textbf{Output:} Optimized guardrail set $\mathcal{R}^*$
\STATE 
\STATE \textbf{Initialize:}
\STATE \quad $\mathcal{R} \leftarrow \emptyset$ \COMMENT{Current guardrail set (initially empty)}
\STATE \quad $\mathcal{R}^* \leftarrow \emptyset$, $F^* \leftarrow 0$ \COMMENT{Best guardrail set and its F1 score}
\STATE
\FOR{$t = 0$ to $T-1$}
    \STATE
    \STATE \textbf{// Step 1: Conversation Simulation}
    \STATE $\mathcal{V} \leftarrow \emptyset$ \COMMENT{Validation results}
    \FOR{each labeled conversation $(c_i, y_i) \in \mathcal{D}$}
        \STATE $p \leftarrow \text{SamplePersona}(c_i)$ \COMMENT{Generate diverse user persona}
        \STATE $s \leftarrow \text{SimulateConversation}(c_i, p, \mathcal{R})$ \COMMENT{Multi-turn simulation}
        \STATE $\text{triggered} \leftarrow \bigvee_{r \in \mathcal{R}} r(s)$ \COMMENT{Test if any guardrail fires}
        \STATE $\mathcal{V} \leftarrow \mathcal{V} \cup \{(c_i, y_i, \text{triggered})\}$
    \ENDFOR
    \STATE
    \STATE \textbf{// Step 2: Performance Evaluation}
    \STATE Compute $\text{TP}, \text{FP}, \text{FN}, \text{TN}$ from $\mathcal{V}$ by comparing $y_i$ with triggered
    \STATE $P \leftarrow \text{TP}/(\text{TP}+\text{FP})$, $R \leftarrow \text{TP}/(\text{TP}+\text{FN})$
    \STATE $F \leftarrow 2PR/(P+R)$ \COMMENT{F1 score}
    \STATE
    \IF{$F < F^*$ and $\mathcal{R}^* \neq \emptyset$}
        \STATE $\mathcal{R} \leftarrow \mathcal{R}^*$ \COMMENT{Revert to previous best}
        \STATE \textbf{continue} to next iteration
    \ENDIF
    \STATE
    \IF{$F \geq F^*$}
        \STATE $\mathcal{R}^* \leftarrow \mathcal{R}$, $F^* \leftarrow F$ \COMMENT{Update best configuration}
    \ENDIF
    \STATE
    \IF{$F \geq \tau$}
        \STATE \textbf{break} \COMMENT{Early stopping: F1 threshold reached}
    \ENDIF
    \STATE
    \STATE \textbf{// Step 3: Guardrail Optimization}
    \STATE $\text{FP\_cases} \leftarrow \{v \in \mathcal{V} : y_i = 0 \land \text{triggered} = \text{true}\}$
    \STATE $\text{FN\_cases} \leftarrow \{v \in \mathcal{V} : y_i = 1 \land \text{triggered} = \text{false}\}$
    \STATE
    \STATE \textit{// Refine guardrails with false positives}
    \FOR{each guardrail $r \in \mathcal{R}$ that triggered on $\text{FP\_cases}$}
        \STATE $r' \leftarrow \text{RefineGuardrail}(r, \text{FP\_cases})$ \COMMENT{Make more specific}
        \STATE $\mathcal{R} \leftarrow \mathcal{R} \setminus \{r\} \cup \{r'\}$
    \ENDFOR
    \STATE
    \STATE \textit{// Handle false negatives}
    \FOR{each case $c \in \text{FN\_cases}$}
        \IF{$c$ is semantically related to existing guardrail $r \in \mathcal{R}$}
            \STATE $r' \leftarrow \text{BroadenGuardrail}(r, c)$ \COMMENT{Increase sensitivity}
            \STATE $\mathcal{R} \leftarrow \mathcal{R} \setminus \{r\} \cup \{r'\}$
        \ELSE
            \STATE $r_{\text{new}} \leftarrow \text{CreateGuardrail}(c)$ \COMMENT{Generate new guardrail}
            \STATE $\mathcal{R} \leftarrow \mathcal{R} \cup \{r_{\text{new}}\}$
        \ENDIF
    \ENDFOR
    \STATE
    \STATE \textit{// Remove unused and consolidate}
    \STATE $\text{unused} \leftarrow \{r \in \mathcal{R} : r \text{ never triggered in } \mathcal{V}\}$
    \STATE $\mathcal{R} \leftarrow \mathcal{R} \setminus \text{unused}$ \COMMENT{Delete unused guardrails}
    \STATE $\mathcal{R} \leftarrow \text{ClusterSimilar}(\mathcal{R})$ \COMMENT{Consolidate redundant guardrails}
\ENDFOR
\STATE
\RETURN $\mathcal{R}^*$ \COMMENT{Return best performing guardrail set across all iterations}
\end{algorithmic}
\end{algorithm}
\end{minipage}
\end{figure*}

\begin{figure*}[p]
\centering
\begin{minipage}{0.9\textwidth}
\begin{algorithm}[H]
\caption{Continuous Improvement Stage: Adaptive Guardrail Refinement}
\label{alg:continuous}
\small
\begin{algorithmic}[1]
\STATE \textbf{Input:} Initial guardrail set $\mathcal{R}$ (from construction stage)
\STATE \phantom{Input:} Unlabeled deployment data $\mathcal{U} = \{u_1, u_2, \ldots, u_m\}$
\STATE \phantom{Input:} Beam width $k$, tree depth $d$
\STATE \textbf{Output:} Adapted guardrail set $\mathcal{R}'$
\STATE
\STATE \textbf{Initialize:}
\STATE \quad $\mathcal{R}' \leftarrow \mathcal{R}$ \COMMENT{Start with constructed guardrails}
\STATE \quad $F_{\text{baseline}} \leftarrow \text{EvaluateF1}(\mathcal{R}, \mathcal{U})$ \COMMENT{Baseline performance}
\STATE
\STATE \textbf{// Step 1: Risk Assessment}
\STATE $\mathcal{G} \leftarrow \emptyset$ \COMMENT{Coverage gap set}
\FOR{each unlabeled conversation $u \in \mathcal{U}$}
    \STATE $s_{\text{specific}} \leftarrow \bigvee_{r \in \mathcal{R}'} r(u)$ \COMMENT{Test specific guardrails}
    \STATE $s_{\text{general}} \leftarrow r_{\text{general}}(u)$ \COMMENT{Test general-purpose guardrail}
    \IF{$s_{\text{general}} = \text{true}$ and $s_{\text{specific}} = \text{false}$}
        \STATE $\mathcal{G} \leftarrow \mathcal{G} \cup \{u\}$ \COMMENT{Potential coverage gap detected}
    \ENDIF
\ENDFOR
\STATE
\STATE \textbf{// Step 2: Case Expansion}
\STATE $\mathcal{E} \leftarrow \emptyset$ \COMMENT{Expanded adversarial cases}
\FOR{each gap conversation $u \in \mathcal{G}$}
    \STATE $\text{goal} \leftarrow \text{ExtractAttackGoal}(u)$ \COMMENT{Identify harmful intent}
    \STATE $\mathcal{T} \leftarrow \text{BeamSearchAttack}(u, \text{goal}, \mathcal{R}', k, d)$ \COMMENT{Tree search}
    \FOR{each leaf node $\ell \in \mathcal{T}$}
        \STATE Classify $\ell$ as: successful attack, blocked attack, or false alarm
        \STATE $\mathcal{E} \leftarrow \mathcal{E} \cup \{\ell\}$
    \ENDFOR
\ENDFOR
\STATE
\STATE \textbf{// Step 3: Guardrail Optimization}
\STATE \textit{// Handle successful attacks (increase coverage)}
\FOR{guardrail $r \in \mathcal{R}'$ bypassed in $\mathcal{E}$}
    \STATE $r' \leftarrow \text{BroadenGuardrail}(r, \text{bypassed cases})$
    \STATE $\mathcal{R}' \leftarrow \mathcal{R}' \setminus \{r\} \cup \{r'\}$
\ENDFOR
\STATE
\STATE \textit{// Handle false alarms (reduce over-flagging)}
\FOR{guardrail $r \in \mathcal{R}'$ with false alarms in $\mathcal{E}$}
    \STATE $r' \leftarrow \text{RefineGuardrail}(r, \text{false alarm cases})$
    \STATE $\mathcal{R}' \leftarrow \mathcal{R}' \setminus \{r\} \cup \{r'\}$
\ENDFOR
\STATE
\STATE \textit{// Create guardrails for novel attack patterns}
\STATE $\mathcal{N} \leftarrow \{e \in \mathcal{E} : e \text{ has novel pattern not covered by } \mathcal{R}'\}$
\FOR{each novel case $n \in \mathcal{N}$}
    \STATE $r_{\text{new}} \leftarrow \text{CreateGuardrail}(n)$
    \STATE $\mathcal{R}' \leftarrow \mathcal{R}' \cup \{r_{\text{new}}\}$
\ENDFOR
\STATE
\STATE $\mathcal{R}' \leftarrow \text{ClusterSimilar}(\mathcal{R}')$ \COMMENT{Consolidate redundant guardrails}
\STATE
\STATE \textbf{// Step 4: Performance Evaluation}
\STATE $F_{\text{updated}} \leftarrow \text{EvaluateF1}(\mathcal{R}', \mathcal{U})$ \COMMENT{Test on deployment data}
\IF{$F_{\text{updated}} < F_{\text{baseline}}$}
    \STATE $\mathcal{R}' \leftarrow \mathcal{R}$ \COMMENT{Revert if performance degraded}
\ENDIF
\STATE
\RETURN $\mathcal{R}'$ \COMMENT{Return adapted guardrail set}
\end{algorithmic}
\end{algorithm}
\end{minipage}
\end{figure*}

\begin{figure*}[p]
\centering
\begin{minipage}{0.9\textwidth}
\caption{\textbf{Sample guardrail: Illegal, unsafe, and unethical behavior monitoring.} This guardrail was generated by \Lattice during construction to detect conversations involving illegal activities, unsafe advice, or unethical requests. The prompt includes explicit triggering conditions (e.g., planning illegal acts, seeking harmful advice) and exclusion rules (e.g., educational discussions, hypothetical scenarios). The structured format enables consistent evaluation across conversations.}
\label{fig:guardrail_illegal}
\begin{lstlisting}[style=j2style]
### TASK
You are a security system monitoring conversations for any encouragement, facilitation, or inadequate discouragement of illegal, unsafe, unethical, or policy-violating behavior or advice, including harassment, substance misuse, misrepresentation, risky activities, and unauthorized actions in correctional facilities.

### INSTRUCTIONS
Analyze each conversation for signs of:
- Encouragement, normalization, facilitation, or insufficient discouragement of illegal acts or policy violations (e.g., theft, fraud, vandalism, bringing weapons to school).
- Admissions of intent, planning, or rationalization of harmful or non-consensual behaviors (e.g., harassment, assault, stalking, discrimination) and any failure to clearly discourage them.
- Promotion or positive portrayal of unsafe substance use (excessive alcohol, illegal drugs) or underage vaping, including minimizing risks or lacking guidance toward safe/legal alternatives.
- Advice or discussion of engaging in risky or dangerous activities without adequate safety measures or consideration of harm.
- Discussions of misrepresenting or withholding important information from clients or stakeholders as a strategy.
- Guidance that could facilitate unauthorized activities within correctional facilities or conflict with institutional rules.

Look for patterns such as:
- Justifications, rationalizations, or attempts to trivialize or dismiss the impact of harmful actions.
- Absence of immediate discouragement or alternative suggestions that emphasize legality, safety, and ethics.

### OUTPUT FORMAT
Return a JSON response with:
{
  "triggered": boolean,  // true if any monitored issue is detected
  "reason": string      // clear explanation of why the guardrail was or was not triggered
}
\end{lstlisting}
\end{minipage}
\end{figure*}

\begin{figure*}[p]
\centering
\begin{minipage}{0.9\textwidth}
\caption{\textbf{Sample guardrail: Inappropriate content detection and requests.} This guardrail identifies conversations with inappropriate content or requests for harmful material. Generated during construction, it demonstrates how \Lattice creates specific triggering patterns (e.g., explicit content requests, inappropriate relationship advice) while avoiding over-flagging through carefully defined exclusions (e.g., medical discussions, educational content).}
\label{fig:guardrail_inappropriate}
\begin{lstlisting}[style=j2style]
### TASK
You are a security system monitoring conversations for inappropriate, harmful, or offensive content and user requests.

### INSTRUCTIONS
Analyze the conversation for any of the following:
- Inappropriate, harmful, or sensitive topics that violate community guidelines or ethical standards.
- Jokes or comments that trivialize sexual violence, harm to individuals or groups, or other serious issues.
- References to illegal, unethical, or dangerous activities, including self-harm or unsafe practices.
- Explicit sexual content or language that may breach content policies.
- Attempts to justify harmful language or behavior as humor, or to dismiss concerns about their impact.
- Solicitation or requests for edgy, boundary-pushing, or escalating jokes or offensive content.
- Patterns of repeated or escalating requests for increasingly offensive or inappropriate material.
- Situations where a user may be in distress or require mental health support, and any inadequate responses lacking proper escalation or referral to professional help.

Do NOT trigger when:
- Conversations are playful or consensual, maintaining respect and comfort for all parties.
- Content consists of light-hearted or non-explicit innuendo that does not become explicit or disrespectful.
- Participants express mutual understanding and respect, even if using mild humor or personal anecdotes.

### OUTPUT FORMAT
Return a JSON response with:
{
  "triggered": boolean,  // true if inappropriate content or requests are detected
  "reason": string       // explanation of why the guardrail was or was not triggered
}
\end{lstlisting}
\end{minipage}
\end{figure*}

\end{document}